\documentclass[11pt]{article}
\usepackage{coling2018}
\usepackage{times}
\usepackage{url}
\usepackage{latexsym}
\usepackage{dsfont}
\usepackage{arydshln}
\usepackage{subfigure}
\usepackage{xcolor}
\usepackage{graphicx}
\usepackage{floatrow}
\newfloatcommand{capbtabbox}{table}[][\FBwidth]

\title{Using $J$-$K$-fold Cross Validation to Reduce Variance When Tuning NLP Models}
\author{Henry B. Moss \\
	STOR-i Centre for \\
	Doctoral Training, \\
	Lancaster University \\\And
	David S. Leslie \\
	Department of Mathematics \\
	and Statistics, \\
	Lancaster University \\
	{\tt initial.surname@lancaster.ac.uk} \\\And
	Paul Rayson \\
	School of Computing \\and Communications, \\
	Lancaster University \\}

\date{}

\begin{document}
\maketitle
\begin{abstract}
$K$-fold cross validation (CV) is a popular method for estimating the true performance of machine learning models, allowing model selection and parameter tuning. However, the very process of CV requires random partitioning of the data and so our performance estimates are in fact stochastic,  with variability that can be substantial for natural language processing tasks. We demonstrate that these unstable estimates cannot be relied upon for effective parameter tuning. The resulting tuned parameters are highly sensitive to how our data is partitioned, meaning that we often select sub-optimal parameter choices and have serious reproducibility issues.

Instead, we propose to use the less variable $J$-$K$-fold CV, in which $J$ independent $K$-fold cross validations are used to assess performance. Our main contributions are extending $J$-$K$-fold CV from performance estimation to parameter tuning and investigating how to choose $J$ and $K$. We argue that variability is more important than bias for effective tuning and so advocate lower choices of $K$ than are typically seen in the NLP literature, instead use the saved computation to increase $J$. To demonstrate the generality of our recommendations we investigate a wide range of case-studies: sentiment classification (both general and target-specific), part-of-speech tagging and document classification.

\end{abstract}
\section{Motivation}
\label{Motivation}

%
%
\blfootnote{
	%
	%
	\hspace{-0.65cm}  

	%
	\hspace{-0.5cm}  
	This work is licenced under a Creative Commons 
	Attribution 4.0 International Licence.
	Licence details:
	\url{http://creativecommons.org/licenses/by/4.0/}
	%
	%
}

In recent years, the main focus of the machine learning community has been on model performance. We cannot, however, hope to improve our model's predictive strength without being able to confidently identify small (but genuine) improvements in performance. We require an accurate measurement of how well our model will perform once in practical use, known as prediction or generalisation error; the model's ability to generalise from its training set (see Friedman et al \shortcite{friedman2001elements} for an in depth discussion). The accurate estimation of model performance is crucial for selecting between models and choosing optimal model parameters (tuning).

Estimating prediction error on the same data used to train a model can lead to severe under-estimation of the prediction error and is unwise. Simple alternatives use a random splitting of data into training and testing sets, or training, validation and testing sets, with the model trained using the training set, tuned on the validation set and performance on the testing set used to report the quality of the fitted model. More sophisticated approaches are based on re-sampling and make more efficient use of the data; including bootstrapping \cite{efron1994introduction} and $K$-fold cross validation (CV) \cite{kohavi1995study}. Since bootstrapping has prohibitive computational cost and is prone to underestimating prediction error, the machine learning community have coalesced around CV as the default method of estimating prediction error. For a snapshot into prediction error techniques currently used in NLP, we look at the proceedings from COLING 2016, focusing on papers that include estimation of model performance. While some submissions use the more sophisticated  $K$-fold CV, the majority use a single data split. Between those that use $K$-fold CV, 19 use 10-fold, 14 use 5-fold and a further 2 use 3-fold. 

Each of these estimation methods involves making one or more random partitions of the data. This partitioning means the estimated prediction error is a random quantity, even when conditioned on the data available to the algorithm. This random splitting of the data leads to variation in our prediction estimates, which we define as internal variability. Although this internal variability has been discussed before \cite{jiang2017error,rodriguez2010sensitivity,bengio2004no}, it is poorly understood for which datasets and models this is a problem. Note that we are referring to variation between separate estimations by $K$-fold CV and not variability between the $K$ prediction error estimates that make up a single round of $K$-fold CV. The estimates are also biased (with their expected value not equal to the truth). Since the model is only trained on a subset it cannot achieve as high performance as if it had access to all the data. Zhang and Yang \shortcite{zhang2015cross} argue that evaluating performance, model selection and parameter tuning have distinct requirements in terms of the bias and variance of our estimators. In particular, as long as bias is approximately constant across different models/parameters, it will have little effect on the selected model. However, variability is critical. If our estimates have variance that swamps the real differences in model performance, we cannot tell the difference between genuinely superior parameters and noise. Reducing the internal variance of cross-validation is the main focus of this paper.

\begin{figure}[]
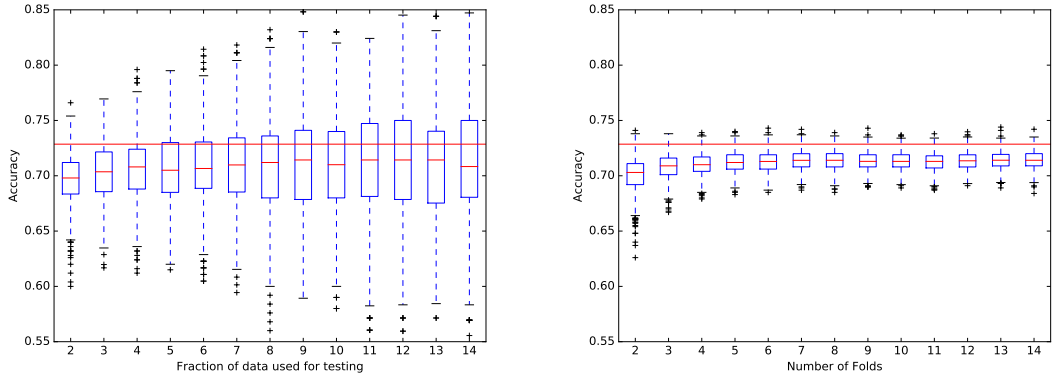

	\begin{center}
		\subfigure[Prediction error estimates based on leaving $\frac{1}{K}$ of the data for testing]{\includegraphics[height=5.5cm]{MOVIES_VAR_Hold_Out_2_labelled}}
		\subfigure[Prediction error estimates based on $K$-fold CV for different choices of $K$]{\includegraphics[height=5.5cm]{MOVIES_VAR_K_2_labelled}}
		\vspace*{-0.5cm}
		\caption{The substantial variability in performance estimates for IMDB sentiment classification. The horizontal line represents the true performance of the model. Each box-plot summarises 1,000 estimations of prediction error based on a different random shuffle of our $1,000$ reviews.}
		\label{MOTIVATION}
	\end{center}
\end{figure}

To motivate the need for the paper and demonstrate the typical size of this variability, we now introduce a simple NLP task; classifying the sentiment of IMDB movie reviews using a random forest (described in detail in Section \ref{prediction}). Figure \ref{MOTIVATION} shows that for  $K$-fold CV and, to an even greater extent single train-test splits, the prediction error estimates have variability substantially larger than performance differences (less than 1\% accuracy improvements between models) identified using the same prediction error estimation methods at COLING 2016. Without analysing the variability of each prediction error estimate it is impossible to say whether we have a genuinely improved model or are just looking at noise. Just because one model outperforms another by a small amount on a particular data-split there is no guarantee that it is genuinely superior and, by changing the partitioning, we may in fact see the opposite. Despite the potentially large variability of the performance estimate, it is common to ignore  the stochasticity and just use the results from a single partitioning. The following arguments and experiments are all with respect to the internal variability of $K$-fold CV. However, as single train-test splits produce less stable prediction error estimates, our arguments are still valid for single data-splits but to an even greater degree.

In the machine learning literature the usual method for reducing the internal variability of prediction error estimates is stratification \cite{kohavi1995study}, used five times at COLING 2016. This keeps the proportions of classes constant across partitioning and so forces each partition to be more representative of the whole data. Stratification reduces variability due to unbalanced classes. However, as demonstrated in Figure \ref{MOTIVATION}, NLP tasks still suffer from large variability even when the classification classes are perfectly balanced. For natural language, representativity is a much more complicated concept than just matching class proportions. Although having been discussed in the corpus linguistics literature \cite{biber1993representativeness}, representativity remains a sufficiently abstract task to not have a clear definition that can be used in NLP. We wish to split our data in a way that avoids under-representing the diversity and variability seen in the whole data, however, complicated structural properties like discourse and syntactic rules make it very difficult to measure this variability.

We present, to the best of the authors' knowledge, the first investigation into prediction error variability specifically for NLP and the first in the machine learning literature to investigate the effect of this variability on parameter tuning, questioning both reproducibility and the ability to reliably select the best parameter value. We argue that variability in our prediction error estimates is often significantly larger than the small performance gains searched for by the NLP community, so these estimates are unsuitable for model selection and parameter tuning (Section \ref{prediction}). We instead propose using repeated $K$-fold CV \cite{kohavi1995study} to reduce this variability (Section \ref{Repeated}). This is not a new idea, yet has failed to be taken up by the NLP community. Repeated CV was used only twice at COLING 2016 and both times only used for improving the stability of prediction error estimates for a chosen model: 10 repetitions of 10-fold CV by Bhatia \shortcite{bhatia2016automatic} and 2 repetitions of 5-fold CV by Collel \shortcite{collell2016image}. Our main contribution is to extend this method to parameter tuning, alongside investigating guidelines for choosing the number of repetitions and $K$. Finally, we demonstrate that stable prediction error estimates lead to more effective parameter tuning across a wide range of typical NLP tasks (Section \ref{Experiments}).

%

\section{Current Practice: Tuning by $K$-fold CV}
\label{prediction}
$K$-fold CV consists of averaging the prediction estimates of $K$ train-test splits, specifically chosen such that each data point is only used in a single test set (for more information see Kohavi \shortcite{kohavi1995study}). The data is randomly split into $K$ folds of roughly equal size (known as  partitioning the data) before each fold is held out to evaluate a model trained on the remaining $K-1$ folds. Increasing $K$ improves stability by averaging over more models. However, each evaluation is performed on a smaller subset of the available data and so the evaluations themselves become less stable. The combination of these competing variabilities makes up the total internal variance which, as confirmed by Figure \ref{MOTIVATION}, is nevertheless smaller than with single train-test splits. If we have enough data that using $1/K^{th}$ of the data still provides stable evaluations then we can see a reduction in internal variability as we increase $K$. However, this is not a general statement \cite{bengio2004no}, as we will demonstrate in Section \ref{Experiments}. It is common to choose  $K=5$ or $K=10$, based on a study by Kohavi \shortcite{kohavi1995study}  where empirical evidence is presented for these choices producing a reasonable trade-off between bias and variance for some specific statistical examples. We will see that the optimal choice  of $K$ is problem-specific so there is no guarantee that Kohavi's studies are comparable to modern NLP tasks. As $K$-fold CV requires the training of $K$ models, there is a clear incentive to choose the smallest suitable value of $K$. This idea is developed in Section \ref{Repeated}.

In this paper, we consider an exhaustive method used to tune parameters; grid search. Here we calculate $K$-fold CV performance estimation across a set of possible parameter values (our grid) and select the value that gives us the best estimated performance. Other popular tuning procedures from the machine learning literature include random search \cite{bergstra2012random} and Bayesian optimisation \cite{snoek2012practical}. Grid search is often not the most efficient approach as it scales poorly with dimensionality. However, its simplicity and interpretability means that it is the standard tuner in NLP and a simple scenario for clearly demonstrating the impact of internal variability on the effectiveness of parameter tuning. Note that all tuning procedures rely on individual prediction error estimates, so our analysis of the problems of grid search is indicative of similar problems with more sophisticated tuning procedures.

To demonstrate the unsuitability of $K$-fold CV for tuning a typical NLP task, we train a sentiment classifier on a bag-of-word features model using a random forest \cite{breiman2001random}, a common set-up in the NLP literature \cite{gokulakrishnan2012opinion,da2014tweet,fang2015sentiment}. We use a large corpus of $25,000$ positive and $25,000$ negative IMDB movie reviews (originally used to train word embeddings \cite{maas2011learning}) and randomly choose a fixed $1,000$ to act as our training set (on which we need to train and tune our model). The purpose of this contrived task is simply to demonstrate the variability that can result from the random partitioning in $K$-fold CV. Thus ``true performance'', consists of the global score on the held-out reviews. We acknowledge that the exact scores depend on which $1,000$ reviews are used for training, so should not be taken as the ground truth. They do, however, allow us to roughly quantify the sub-optimality of our tuned parameters and so measure the effectiveness of different tuning procedures. Even without these scores, the variability of the tuned parameter values raises concerns regarding reproducibility (as decisions cannot be reproduced without exact knowledge of the partitioning).

Using Python's \footnote{\label{SKLEARN} \url{http://scikit-learn.org/stable/index.html}} random forest implementation, we consider tuning the maximum proportion of features ($max\_features$) considered at a single time by our model; a parameter whose accurate tuning is crucial to prevent over-fitting. We initially focus on $K=5$, as Figure \ref{MOTIVATION} shows a lack of significantly improved bias or variance  for the increased computational cost of higher $K$. For each of 1,000 random $5$-fold partitions of our training set, we estimate the performance for each parameter value in the set $\{0.01,0.02,...,0.5\}$ and choose the value that gives us the best performance estimate. Aside from maximum depth and number of trees, which we set to 4 and 100 respectively, we keep all other parameters as their SKLEARN defaults. We limit ourselves to the $300$ features with the highest tf-idf score \cite{salton1988term}. Note that random forests are stochastic, with a sub-sampling step used to fit the individual tree classifiers. However, as this additional variation cannot be fixed between successive performance estimates (with the sampling being dependent on how our data is partitioned) this should be considered as part of the internal variability.

\begin{figure}[h]
	\begin{center}
		\subfigure[Tuning of $max\_features$  by $5$-fold CV.]{\includegraphics[height=5.25cm]{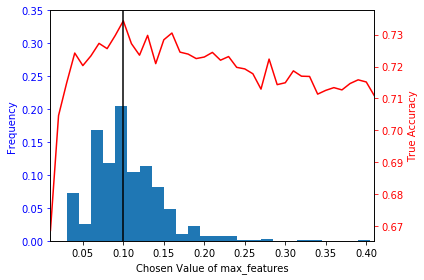}}
		\subfigure[Tuning of $max\_features$ by $10$-$5$-fold CV.]{\includegraphics[height=5.25cm]{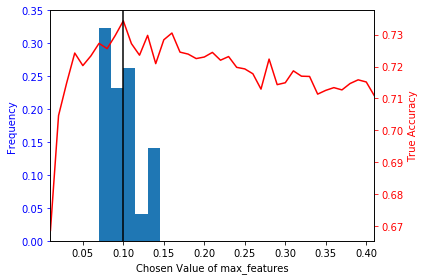}}
		\vspace*{-0.5cm}
		\caption{ Distribution of our chosen $max\_features$ across $1,000$ random data partitions. We compare the industry standard (a) with our proposed solution (b). The global performance of a model for a given value of $max\_features$ is superimposed in red, with the global optimal choice represented by the vertical line.}
		\label{CVIMDB}
	\end{center}
\end{figure}

Figure \ref{CVIMDB}(a) shows that partitioning can have a large effect on the chosen parameter, leading to choices of $max\_features$ across the wide range of $0.03$ to $0.4$ (almost covering the whole search space). This variability means that our tuning procedure is often failing to choose the parameter that gives the best global performance and so produces sub-optimal models. Depending on the initial partitioning, our tuning procedure can lead to tuned models with global performance varying from $71\%$ to $73.5\%$ accuracy. In Section \ref{Experiments}, we will see that the variability is even larger when tuning multiple parameters at once.   

\section{Proposed Approach: Tuning By Repeated $J$-$K$-fold Cross Validation}
\label{Repeated}

Unfortunately, only considering a single partitioning cannot give us information about the amount of variability present in our performance estimates. To produce Figure \ref{CVIMDB}, we had to look at the tuned model resulting from $1,000$ different partition choices. This observation motivates the use of repeated $K$-fold CV, also known as $J$-$K$-fold CV; averaging the $K$-fold CV estimate from $J$ different partition choices. It has been shown empirically that repeated CV reduces internal variability and so stabilises prediction error \cite{chen2012sensitivity,vanwinckelen2015look,jiang2017error}, especially for smaller datasets \cite{rodriguez2010sensitivity}. These authors, however, only investigate the use of $J$-$K$-fold CV to reduce the variability of individual performance estimates and do not consider its use for tuning.

We propose the following novel extension to grid search, which extends the improved stability of $J$-$K$-fold CV to parameter tuning. For each of $J$ random partitionings, we calculate a $K$-fold CV estimate for every parameter choice on our grid. The average over these partitions produces performance estimates for each parameter choice and we choose the parameter value that maximises these stable estimates of prediction error. For the rest of the paper we will refer to this procedure as tuning by $J$-$K$-fold CV. We believe that using $J$-$K$-fold CV will also improve the effectiveness of the more sophisticated parameter tuning procedures, however, this requires further investigation.

Returning to our IMDB example, Figure \ref{CVIMDB}(b) shows that using information from multiple partitioning choices greatly reduces variability in the chosen parameter value, with standard deviation dropping from  $0.0427$ to $0.0221$ when tuning by $10$-$5$-fold CV (Figure \ref{CVIMDB}(b)) instead of vanilla $5$-fold CV (Figure \ref{CVIMDB}(a)). This corresponds to the worse performing tuned model over 1,000 random partition choices improving from  $71.3\%$ to $72.1\%$ accuracy. We also see a significant reduction in the standard deviation of our performance estimate of the tuned model from $0.700\%$ to $0.233\%$, greatly improving our ability to discern between closely performing models. 

\begin{figure}[h]
	\begin{center}
		\subfigure[Variability of the tuned $max\_features$  value.]{\includegraphics[height=4.7cm]{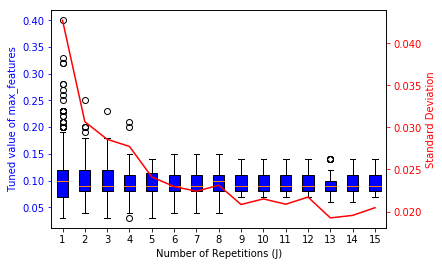}}
		\subfigure[Variability of the performance estimate for the tuned model.]{\includegraphics[height=4.7cm]{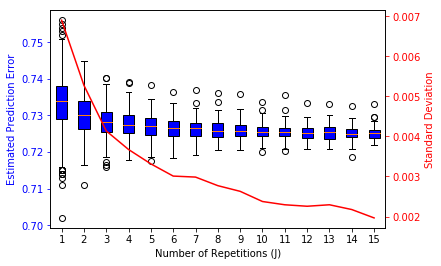}}
		\vspace*{-0.5cm}
		\caption{ Diminishing reductions in variance as we increase the number of repetitions $J$ for tuning by $J$-$5$-fold CV. As before, for each value of $J$ we retune our model  over $1,000$ different random partitioning.}
		\label{Vars}
	\end{center}
\end{figure}

\begin{table}[h]
	\centering
	\scalebox{0.9}{
		\begin{tabular}{llll}
			\hline
			& SD of chosen  & Range of chosen  & SD of the accuracy  \\ & $max\_features$ value&$max\_features$ value& estimate of tuned model \\ \hline
			1-10-fold CV &    0.0396                       &   0.03-0.38                                &      0.550  \%                                       \\
			2-5-fold CV  &   0.0307                          &   0.04-0.25                               &  0.528    \%                                          \\ \hdashline
			1-20-fold CV  &    0.0337                       &   0.03-0.24                               &          0.483 \%                                     \\
			2-10-fold CV &     0.0301                         &  0.03-0.23                                & 0.400    \%                                           \\
			4-5-fold CV  &       0.0278                      & 0.03-0.21
			&     0.367  \%                                         \\
			\hline
	\end{tabular}}
	\caption{Standard deviation (SD) performance of $J$-$K$-fold tuning on the IMDB dataset over $1,000$ different partitions (3 s.f).}
	\label{Jdemo}
\end{table}

We now discuss how to choose the values of $J$ and $K$ in terms of computational cost and effective tuning, a concept we define as selecting near-optimal parameter values irrespective of how the data is partitioned. At a first approximation the cost of $J$-$K$-fold CV is proportional to $J*K$ (the number of models trained and evaluated). If, as is the case for sophisticated models, training and evaluation costs are more than linear in the amount of data, then the computational cost grows faster in $K$, but are still linear in $J$. So for fixed computational resources we have a trade-off between increasing either $J$ or $K$.

First, consider increasing $K$. As previously mentioned there is no clear general relationship between $K$ and the internal variability of $K$-fold CV. As long as $K\geq3$ (to guarantee overlapping training sets), we can have comparable variance across $K$ (see Section \ref{Experiments}). This means that the only reliable reason to increase $K$ is to reduce bias, however, there is no clear argument for how the bias of each individual performance estimate relates directly to bias in our tuning procedure (the difference between the expected value of our tuned parameter and the true optimal parameter choice). As long as the bias of the estimate for each parameter choice is roughly constant across the parameter grid, then we would expect this bias to have a limited effect on our tuned parameter value (as is assumed for tuning by vanilla $K$-fold CV). A further research interest is to investigate exactly when this assumption is violated and the effect that this can have on the effectiveness of parameter tuning. 

In contrast, increasing $J$ has no effect on bias but does significantly reduce the internal variability. For fixed data, the estimations from each repetition are independent and identically distributed, so prediction error estimates from vanilla $K$-fold CV have $J$ times as much variability as $J$-$K$-fold CV but the same bias. Figure \ref{Vars} demonstrates that we see similar variance reduction returns in the choice from tuning and also the performance estimate of the tuned model as we increase $J$. This means we can access the largest drops in variability within the first few repetitions \cite{kim2009estimating}.

We can therefore consider our choices of $K$ and $J$ in isolation. $K$ is increased to reduce bias whereas $J$ reduces the internal variability. We commented earlier that effective parameter tuning is more sensitive to variance than bias. This is confirmed by our IMDB task, where we see that although the tuned value from $5$-fold CV has a mean close to the best performing parameter choice (calculated for our held-out $49,000$ reviews), the tuned value has significant variability. Sub-optimal tuning is  a consequence of this large variance, which overshadows any potential problems from the bias. Therefore we first need to reduce internal variability before  our tuning can benefit from reduced bias.

Table \ref{Jdemo} summarises the tuning performance of five different $J$ and $K$ choices. We see that between $1$-$10$-fold CV and $2$-$5$-fold CV, and between $2$-$10$-fold, $4$-$5$-fold and $1$-$20$-fold (choices of equivalent computational cost), we have the least variability when reducing $K$ and using the saved computation for increasing $J$. Note that the naive approach of using all available computation to increase $K$ ($1$-$20$-fold CV) produces a wider range of parameter choices than $2$-$10$-fold and $4$-$5$-fold CV. This is despite any bias reductions from choosing a higher $K$ and so we do not recommend $1$-$J*K$-fold CV. This analysis questions the rule of thumb selection of $K=10$ common in NLP, as $2$-$5$ CV seems to produce superior tuning at the same cost, a hypothesis we now test across a range of NLP tasks.

\section{Case Studies}
\label{Experiments}

We now widen our investigation into the suitability of $K$-fold CV by analysing three further NLP tasks, chosen to cover a wide range of typical models and datasets. Note that we are not trying to improve the modelling of these standard models, just showing that ineffective tuning due internal variability is a general issue across NLP. We look at a simple part-of-speech tagger using logistic regression, document classification with support vector machines, and target-dependent sentiment classification with an LSTM. Each task is treated the same as our IMDB example; comparing the performances of tuning by $1$-$10$-fold CV (the industry standard) against the computationally equivalent $2$-$5$-fold CV. We also investigate the situation where we have two and four times the computational budget by comparing $2$-$10$-fold with $4$-$5$-fold and $4$-$10$-fold with $8$-$5$-fold. Our first task shows a case where tuning by vanilla $K$-fold is still reasonably effective. For task two we tune multiple parameters at once, seeing substantial variation in our tuning and so a real need for $J$-$K$-fold CV. Our final example shows that our criticisms still hold for more sophisticated models, where it can be necessary to have even more than $8$ repetitions for reliable tuning. All code is available \footnote{\label{CODE} \url{https://github.com/henrymoss/COLING2018}}.

\begin{figure}[h]
	\begin{center}
		\subfigure[$K$-fold CV for logistic regression part-of-speech tagger.]{\includegraphics[height=3.25cm]{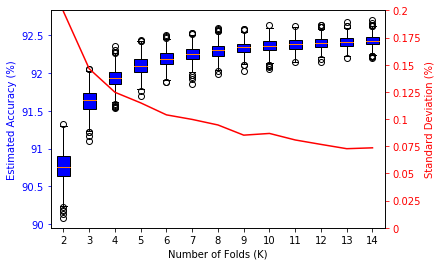}}
		\subfigure[$K$-fold CV for support vector machine topic classifier ]{\includegraphics[height=3.25cm]{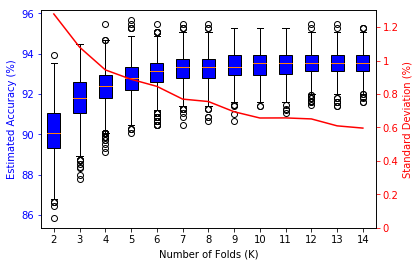}}
		\subfigure[$K$-fold CV for LSTM target-dependent sentiment classification. ]{\includegraphics[height=3.25cm]{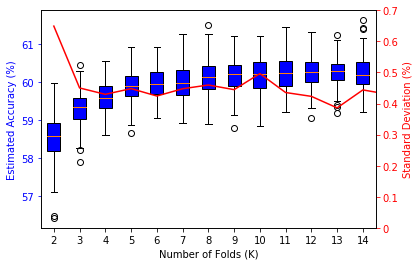}}
		\caption{Prediction error estimates based on $K$-fold CV. The variation is calculated across $1000$ random partitionings for (a) and (b)) and $100$ for (c).}
		\label{KVAR}
	\end{center}
\end{figure}

Before investigating tuning the different models, we use these examples to provide empirical evidence for our recommendation of using lower $K$ than the common choice of ten. We argued that increasing $K$ has no clear effect on the internal variance of individual performance estimates from $K$-fold CV and diminishing bias reduction returns. These are presented in Figure \ref{KVAR}. Note that although our POS tagger and topic classifier do in fact become less variable for larger $K$, this is not a general rule, with our LSTM providing a counter-example (with comparable variance for choices of $K$ above three). Even when we do see reductions in variability when increasing $K$ beyond five, the following experiments show that reductions in internal variability from increasing $J$ dominate those gained from higher $K$, failing to provide justification for choosing $K$ as large as ten. Note also that the accuracy estimates for all three tasks show significant variation and so predictions by vanilla $K$-fold CV are unsuitable for the comparison of even relatively closely performing models.

\subsection{Part-of-speech tagger}
\label{Example1}

For our first task we are not directly recreating a method from a particular paper, just demonstrating that $J$-$K$-fold CV can improve the tuning of even the most common NLP models. To find such a task we consulted \newcite{jurafsky2000speech}, choosing logistic regression and part-of-speech tagging. We train and evaluate our model on a fixed $10,000$ word subset of the Brown Corpus (as available in Python's NLTK package \cite{bird2006nltk}). As with our IMDB example, we use the remaining $90,000$ words as an independent global test set to check the validity of our parameter tuning. We classify with respect to the Penn Treebank tagging guidelines \cite{santorini1990part} based on simple intuitive features including information about prefixes, suffixes, capital letters, position in sentence and adjacent words. We tune the amount of l2 regularisation (described by $C$ in SKLEARN) on the grid $\{0.001,0.005,0.01,0.05,0.1,0.5,1,5,10,50,100\}$. 

Table \ref{POS} shows that larger choices of $J$ with smaller $K$ provide the most effective tuning at each computational budget. Our most unstable tuning is by $1$-$10$-fold CV. However when checked on the held-out population, the tuned model with the worst true performance only loses $0.05\%$ accuracy from the parameter choice with best global performance. So although increasing $J$ does reduce variability, it is not always necessary for effective tuning, as vanilla $10$-fold CV produced consistently near-optimal models. The size of variability, however, is still a concern for reproducibility, as different random partitions can lead to choosing $C$ as $10$, $50$ or $100$.

\begin{figure}
	\begin{floatrow}
		
		\capbtabbox{%
			\scalebox{0.9}{
			\begin{tabular}{lll}
				\hline
				& SD of  & SD of the \\
				&chosen&accuracy estimate \\ 
				&$C$ value         & of the tuned model\\ \hline
				1-10-fold CV &   34.6                                       &   0.0915  \%                                          \\
				2-5-fold CV  &   31.0                                         &       0.0874   \%                                     \\\hdashline
				2-10-fold CV &   29.0                                         &       0.0653   \%                                     \\
				4-5-fold CV  &   27.3                                         &       0.0653   \%                                     \\\hdashline
				4-10-fold CV &   23.2                                         &       0.0472    \%                                    \\
				8-5-fold CV  &   24.4                                         &  0.0439     \%                                        \\ \hline
			\end{tabular}}

		}{%
			\caption{Performance of different $J$-$K$-fold tuning procedures for a logistic regression part-of-speech tagger over $1,000$ different partitions (3 s.f).}
			\label{POS}
		}
		\capbtabbox{%
			\scalebox{0.9}{
			\begin{tabular}{llll}
				\hline
				SD of & SD of & SD of the \\chosen&chosen& accuracy estimate\\ 
				$C$ value   & $gamma$ value   & of the tuned model \\ \hline
				28.7                           & 0.121                                 &  0.807            \%                                 \\
				30.7                             & 0.0987                                 &    0.615          \%                                 \\ \hdashline
				21.3                            &  0.105                                &   0.580              \%                              \\
				20.4                            & 0.0766                                 &     0.445        \%                                  \\ \hdashline
				20.0                           &  0.0708                                &    0.407         \%                                 \\
				17.5                             &  0.0563                                &    0.330        \%                                   \\ \hline
			\end{tabular}}
			
		}{%
			\caption{Performance of different $J$-$K$-fold tuning procedures for a support vector machine topic-classifier (3 s.f).}
			\label{TOPIC}
		}
	\end{floatrow}
\end{figure}

\subsection{Document Classification}
\label{Example2}
We now consider a task where tuning by vanilla $K$-fold CV is inadvisable; topic classification on the Reuters-21578 dataset \cite{lewis1997reuters21578} via support vector machines (SVM) \cite{cortes1995support}. This exact task is well-studied in the NLP literature \cite{joachims1998text,tong2001support,leopold2002text} however no details are provided regarding the tuning of model parameters. Note that SVM are still commonly used for document classification. This dataset and a specific train-test split (known as the ApteMod version) are available in NLTK. For ease of explanation, we focus on just the corn and wheat categories, producing a binary classification task with 334 instances. We tune two parameters; the flexibility of the decision boundary and the RBF kernel coefficient (denoted in SKLEARN as $C$ and $gamma$ respectively).  We search for $C$ in $\{1,5,10,50,100,500,1000,5000,10000\}$ and $gamma$ in $\{0.05,0.1,0.15,0.2,0.25,0.3,0.35,0.4,0.45\}$.

\begin{figure}[h]
	\begin{center}
		\subfigure[Tuning SVM by vanilla $5$-fold CV]{\includegraphics[height=4.5cm]{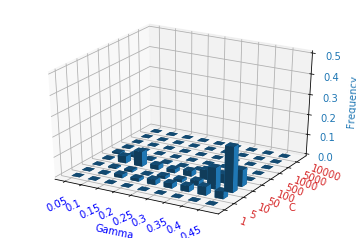}}
		\subfigure[Tuning SVM by $2$-$5$-fold CV]{\includegraphics[height=4.5cm]{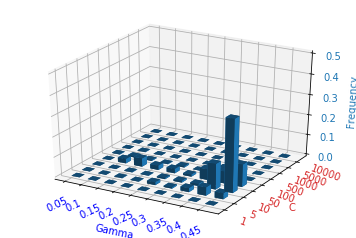}}
		\subfigure[Tuning SVM by $8$-$5$-fold CV]{\includegraphics[height=4.5cm]{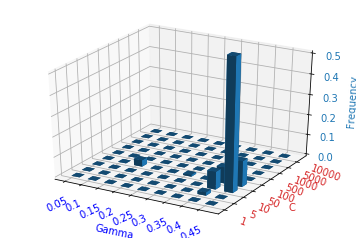}}
		\vspace*{-0.5cm}
		\caption{ Distribution of the tuned parameter values across $1,000$ random partitionings.}
		\label{SVM_2d}
	\end{center}
\end{figure}

As summarised in Table \ref{TOPIC}, this task suffers from much larger variability than the part-of-speech tagger. Once again, at each computational cost, the most effective tuning  (in terms of $gamma$ and variability of the performance estimate of the tuned model) is from the lower choice of $K$ and higher $J$. Note that for $C$ variability, there is an slight increase when moving from $1$-$10$-fold to $2$-$5$-fold CV, however, this corresponds to a large drop in the other variabilities and so $2$-$5$-fold CV still provides more effective tuning overall. Also note that we see a larger reduction in $gamma$ variability than $C$ variability for larger $J$ choices. This is explained in Figure \ref{SVM_2d}, where we see substantially more variation in the tuned value of $gamma$ than in $C$ when tuning with no repetitions (Figure \ref{SVM_2d}(a)). Note that effective parameter tuning corresponds to a single predominately large bin, i.e. selecting a single parameter choice irrespective of the partitioning. This is achieved by increasing $J$ (Figures \ref{SVM_2d}(b) and \ref{SVM_2d}(c)). Also note that the accuracy estimate for the model tuned by $1$-$10$-fold CV is not stable enough to allow comparison with other models of close performance. To be able to reliably distinguish between our tuned model and alternatives with performance differing by only a couple of percentage points, we require higher choices of $J$.

\subsection{Target-Dependent Sentiment Classification}
\label{Example3}
For our final task we consider a much more sophisticated model  using LSTMs (with Python's KERAS \footnote{\label{KERAS} \url{https://keras.io/}}) to perform target-dependent sentiment classification \cite{tang2015effective} using twitter-specific sentiment word vectors \cite{tang2014learning} on the benchmark twitter dataset collected by Dong et al \shortcite{dong2014adaptive}. Our implementation relies heavily upon the reproduction study of Moore et al \shortcite{moore_2018}. Each of $6,248$ sentences are annotated with a target element and the task is to predict the sentiment regarding that element (positive, negative or neutral). Unfortunately little information is provided about model architecture or parameter tuning. However, we can still investigate the effectiveness of tuning by $J$-$K$-fold CV. Throughout these experiments we fix the maximum number of epochs as $100$ and stop training when we see no improvements in validation set performance for five successive epochs. For each of our $J*K$ models, the validation set is a random $20\%$ of that model's training data, and so can be thought of just another part of the random partitioning. We use an ADAM optimiser with the default learning parameters and a batch size of $32$.

\begin{figure}[h]
	\begin{center}
		\subfigure[Tuning width by $1$-$10$-fold CV and $8$-$5$-fold CV]{\includegraphics[height=4.5cm]{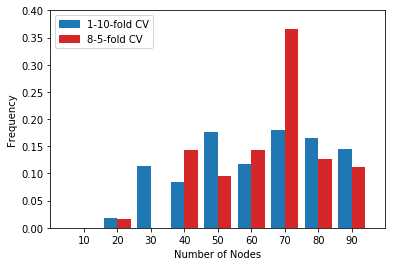}}
		\subfigure[Tuning regularisation by $10$-fold CV]{\includegraphics[height=5cm]{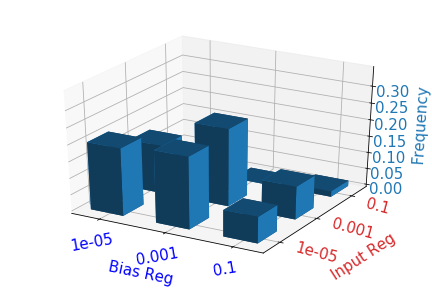}}
		\subfigure[Tuning regularisation by $8$-$5$-fold CV]{\includegraphics[height=5cm]{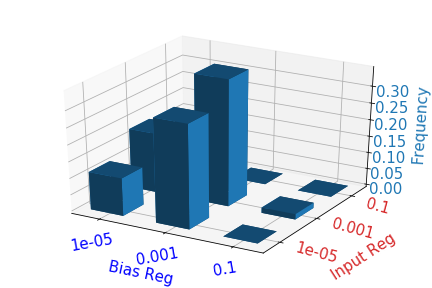}}
		\vspace*{-0.5cm}
		\caption{ Distribution of tuned LSTM values over random partitionings.}
		\label{LSTMFIG}
	\end{center}
\end{figure}

We perform two experiments: tuning the number of nodes in the LSTM layer (known as width) across the grid $\{10, 20, 30,.., 90\}$, and separately tuning the amount of l2 regularisation on the inputs and biases (for a fixed width of $50$) each across $\{0.00001,0.001,0.1\}]$. Tables \ref{LSTM} and \ref{LSTM2} provide further support for our claim that the most effective tuning for a specified computational budget comes from lower $K$ and higher $J$. Note that, as shown in Figure \ref{LSTMFIG}(a), vanilla $1$-$10$-fold CV is not at all suitable for tuning the width of our LSTM, as it produces almost uniform values between $30$ and $90$. It also fails to consistently select a single choice for the regularisation scheme (Figure \ref{LSTMFIG}(b)). In contrast, tuning with $8$-$5$-fold produces much more consistent choices, the majority of the time choosing $70$ for the optimal width (Figure \ref{LSTMFIG}(a)) and $0.001$ for both the input and bias regularisation (Figure \ref{LSTMFIG}(c)). Although variability in our chosen LSTM parameters does reduce as we increase $J$, it remains significantly higher than the gaps in our parameter grid. When coupled with the relative stability of the accuracy estimates of this tuned model, this suggests that the performance differences between the most frequent choices by $8$-$5$-fold CV are small. However, to consistently tune the model to this specificity we require larger choices of $J$ than $8$.

\begin{figure}
	\begin{floatrow}
		
		\capbtabbox{%
			\scalebox{0.9}{
			\begin{tabular}{lll}
				\hline
				& SD  of  & SD of the  \\
				& chosen & accuracy estimate \\ 
				&widths &of the tuned model\\ \hline
				1-10-fold CV &   19.9                                       &   0.293   \%                                         \\
				2-5-fold CV  &   19.8                                         &       0.240   \%                                    \\\hdashline
				2-10-fold CV &   19.3                                         &       0.213   \%                                     \\
				4-5-fold CV  &   18.4                                         &       0.205   \%                                     \\\hdashline
				4-10-fold CV &   17.3                                        &       0.169  \%                                     \\
				8-5-fold CV  &   16.0                                         &  0.166  \%                                           \\ \hline
			\end{tabular}}
		}{%
			\caption{Performance of $J$-$K$-fold tuning  when choosing the width of the LSTM (3 s.f).}%
			\label{LSTM}
		}
		\capbtabbox{%
			\scalebox{0.9}{
			\begin{tabular}{llll}
				\hline
				SD of   & SD of  & SD of the  \\ 
				chosen bias&chosen input& accuracy estimate \\ 
				regularisation & regularisation&of the tuned model \\ \hline
				0.0404                       &   0.0160                               &      0.365  \%                                       \\
				0.0285                        &   0.00705                              &  0.284    \%                                          \\ \hdashline
				0.0358                       &   0.0166                               &          0.254 \%                                     \\
				0.0195                        &  0.00995                                & 0.206    \%                                           \\ \hdashline
				0.0280                    & 0.0185
				&     0.196 \%                                         \\
				0.00815                      & 0.000500
				&     0.148  \%                                         \\
				\hline
			\end{tabular}}
		}{%
			\caption{Performance of $J$-$K$-fold tuning on the IMDB dataset over $1,000$ different partitions (3 s.f).}
			\label{LSTM2}
		}
		
	\end{floatrow}
\end{figure}

\newpage
\section{Discussion and Conclusions}

The aim of this paper has been to demonstrate the significant variability of current performance estimation techniques when applied to NLP tasks. We argue that the size of this variability is poorly understood by practitioners, which has serious implications for both model selection and parameter tuning. We show that the variability in estimates can often be larger than the performance improvements sought by the NLP community; questioning conclusions regarding the comparison of techniques. For parameter tuning, the variability can lead to unstable, irreproducible and significantly sub-optimal parameter choices.  

We advocate the use of $J$-$K$-fold CV, which we extend to parameter tuning. By using information from multiple estimations we stabilise our tuning procedure. To counteract the computational cost of increasing $J$, we suggest lower choices of $K$, as effective tuning is more reliant on variability than bias.

Although we have shown the effectiveness of some specific choices of $J$ and $K$ on our NLP examples, there is still work to be done regarding choosing their optimal configuration, which is problem dependent. Unlike $K$, $J$ can be chosen adaptively, allowing practitioners to respond to the amount of observed variability and efficiently manage computational resources. We would also like to analyse the current common practice of early stopping (as in our LSTM example), which requires evaluations on another held-out set to prevent over-fitting. This is likely to suffer from the concerns outlined in this report and it is poorly understood how best to incorporate early stopping into a wider parameter tuning framework.

\section{Acknowledgements}

The authors are grateful to reviewers, whose comments and advice have greatly improved this paper.  The research was supported by EPSRC and the STOR-i Centre for Doctoral Training.

\newpage

\bibliography{References}

\end{document}